\documentclass[letterpaper, 10 pt, conference]{ieeeconf}  
\usepackage{graphicx}
\usepackage{makecell}
\usepackage{booktabs}
\usepackage{amsmath}

\IEEEoverridecommandlockouts                              

\title{\LARGE \bf
Adjusting Tissue Puncture Omnidirectionally In Situ with Pneumatic Rotatable Biopsy Mechanism and Hierarchical Airflow Management in Tortuous Luminal Pathways
}

\author{Botao Lin, Tinghua Zhang, Sishen Yuan, Tiantian Wang, Jiaole Wang, Wu Yuan, Hongliang Ren$^*$
\thanks{The work was supported by Hong Kong Research Grants Council (RGC) Collaborative Research Fund (CRF C4026-21GF), and General Research Fund (GRF 14203323), Research Impact Fund (RIF R4020-22), Guangdong Basic and Applied Basic Research Foundation (GBABF) \#2021B1515120035.}
\thanks{B. Lin, S. Yuan, H. Ren are with the Department of Electronic Engineering, The Chinese University of Hong Kong, Shatin, Hong Kong, China.}
\thanks{T. Zhang, W. Yuan are with the Department of Biomedical Engineering, The Chinese University of Hong Kong, Shatin, Hong Kong, China.}
\thanks{T. Wang is with the School of Robotics and Advanced Manufacturing, Harbin Institute of Technology, Shenzhen, China.}
\thanks{J. Wang is with the School of Medical and Engineering, Harbin Institute of Technology, Shenzhen, China.}
\thanks{$^*$Corresponding author: Hongliang Ren (e-mail: hlren@ieee.org).}
}

\begin{document}

\maketitle
\thispagestyle{empty}
\pagestyle{empty}

%
%
%
%

\begin{abstract}
\textit{In situ} tissue biopsy with an endoluminal catheter is an efficient approach for disease diagnosis, featuring low invasiveness and few complications.
However, the endoluminal catheter struggles to adjust the biopsy direction by distal endoscope bending or proximal twisting for tissue sampling within the tortuous luminal organs, due to friction-induced hysteresis and narrow spaces.
Here, we propose a pneumatically-driven robotic catheter enabling the adjustment of the sampling direction without twisting the catheter for an accurate \textit{in situ} omnidirectional biopsy.
The distal end of the robotic catheter consists of a pneumatic bending actuator for the catheter's deployment in torturous luminal organs and a pneumatic rotatable biopsy mechanism (PRBM).
By hierarchical airflow control, the PRBM can adjust the biopsy direction under low airflow and deploy the biopsy needle with higher airflow, allowing for rapid omnidirectional sampling of tissue \textit{in situ}.
This paper describes the design, modeling, and characterization of the proposed robotic catheter, including repeated deployment assessments of the biopsy needle, puncture force measurement, and validation via phantom tests.
The PRBM prototype has six sampling directions evenly distributed across 360 degrees when actuated by a positive pressure of 0.3 MPa.
The pneumatically-driven robotic catheter provides a novel biopsy strategy, potentially facilitating \textit{in situ} multidirectional biopsies in tortuous luminal organs with minimum invasiveness.  

\end{abstract}


%
%
%
%

\section{Introduction}
\begin{figure}[t]
    \centering\includegraphics[width=0.45\textwidth]{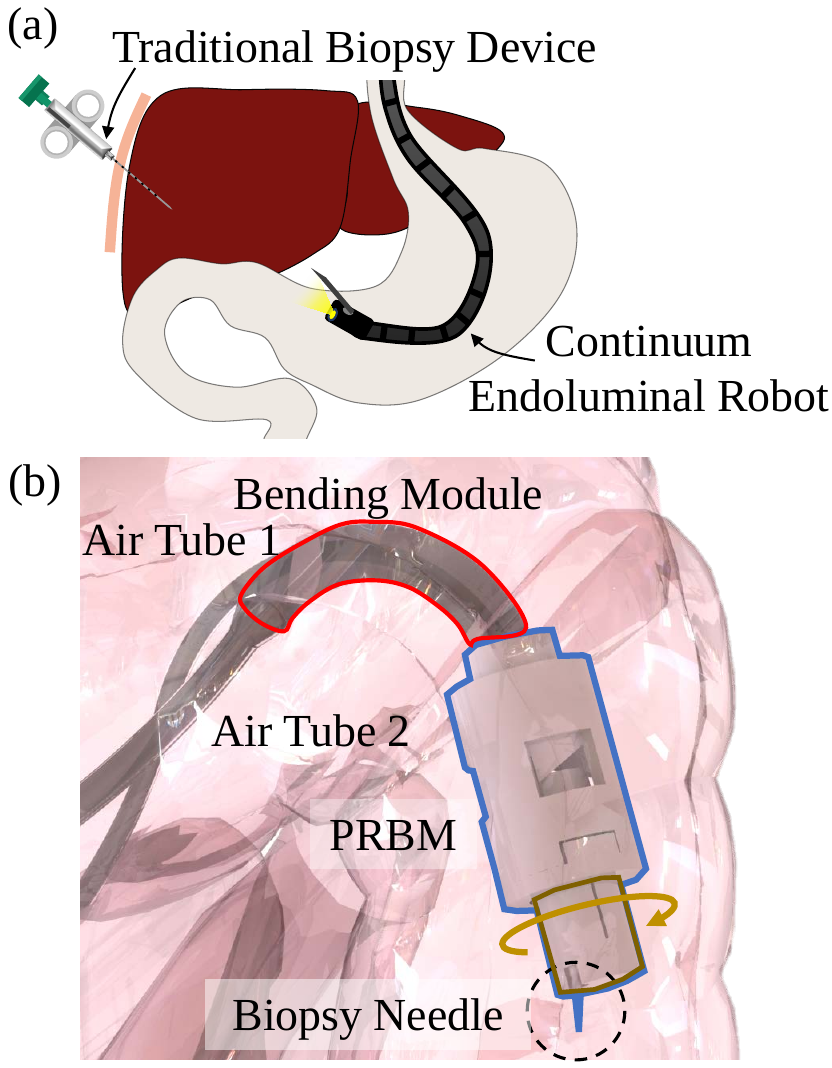}
    \caption{Comparison between the traditional biopsy tool, continuum endoluminal robot, and proposed robotic catheter.
    (a) Traditional hand-held soft tissue biopsy needle and the continuum endoluminal robot with biopsy function.
    (b) Concept view of the proposed robotic catheter.
    With the help of a bending module, the robot can access a cavity deep inside the human body.
    The PRBM mounting on the distal end enables fast variable-directional biopsy, even when the robot goes deep inside the body. }
    \label{fig:showmodules}
\end{figure}

\begin{figure*}[t]
    \centering
    \includegraphics[width = 0.95\textwidth]{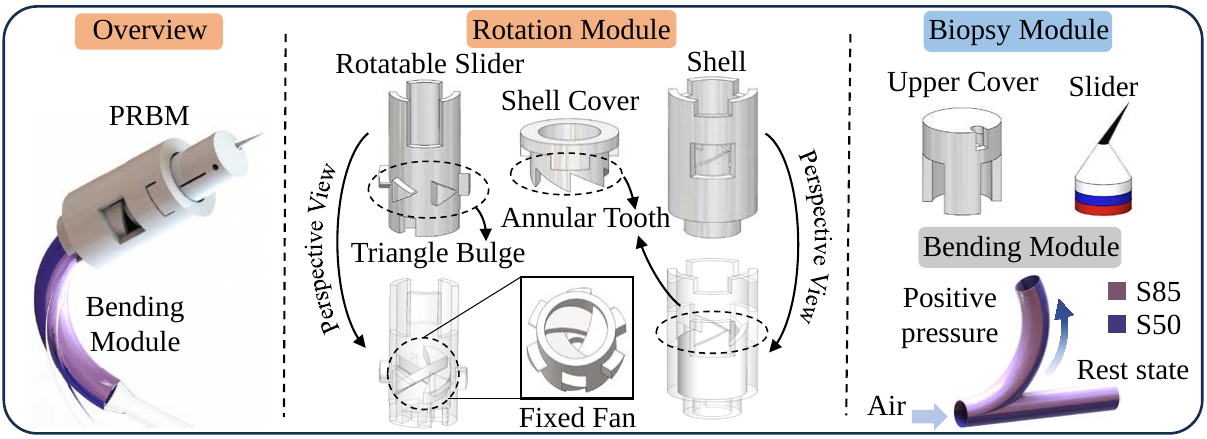}
    \caption{Design of the proposed robotic catheter.
    An overview of the PRBM and its components is shown. 
    The PRBM is composed of a rotation module and a biopsy module.
    The rotation module consists of a rotatable slider, a shell cover, and an outer shell.
    The biopsy module consists of an upper cover and a slider.
    The bending module is composed of two kinds of material with different tensile modulus, i.e., FLXA-CT-\textbf{S50} and FLXA-CT-\textbf{S85}.
    When the positive air pressure is supplied, the bending module would bend toward the side of FLXA-CT-\textbf{S85}.}
    \label{fig:design}
\end{figure*}

\textit{In situ} biopsy has long been the gold standard for disease diagnosis. 
Compared to traditional needle biopsies by piercing organs percutaneously from the outside \cite{strnad2024percutaneous}, endoluminal robotic biopsies, which can employ a catheter to access targeted regions of the human body through tiny incisions or natural cavities, are becoming a prominent research focus due to lower invasiveness and complications \cite{lin2022modular,duan2024survey}. 
For example, Lenny \textit{et al.} \cite{dupourque2019transbronchial} proposed a three-segment tendon-driven bronchial biopsy catheter robot to perform follow-the-leader movement, exhibiting a better steering performance than traditional biopsy catheters.
Wu \textit{et al.} \cite{wu2017development} proposed a concentric tube robot system for efficient posterior nasopharynx biopsy.
Giovanni \textit{et al.} \cite{pittiglio2022patient} designed a soft magnetic catheter, allowing the deployment in the bronchus for tracheal biopsy.
\begin{table*}[t]
\centering
\caption{Comparison between the proposed robot and other biopsy devices}
\begin{tabular}{ccccc}
    \toprule
     & \makecell[c]{Invasiveness to human body} &\makecell[c]{Flexibility of biopsy tool}  & \makecell[c]{Working space of biopsy tool in confined cavity} \\
    \midrule
    Traditional Biopsy Tool                                            & High                           & Low                        & Low                                             \\
     Traditional endoluminal Robot                                            & Low                            & High                       & Normal                                          \\
    Proposed Robotic Catheter                                     & Low                            & High                       & High
                                        \\
    \bottomrule                 
\end{tabular}
\label{tab:benchmark}
\end{table*}
In sum, these robots demonstrated excellent mobility and accessibility in luminal organs for \textit{in situ} biopsies.
However, when the endoluminal robots are deployed into the tortuous luminal organs, it is challenging to control their distal end due to inefficient torque transfer and high friction, which hinders accurate targeted biopsies \cite{russo2023continuum}.

Although pushing, pulling, and twisting the robotic catheter in the proximal end can adjust the direction of biopsy,  extensive contact and collision with the inner surface of the luminal organs cause discomfort to the patients and even injuries and complications \cite{li2023three}.
Additionally, these applied forces increase the robot's torsion, decreasing the stability and reducing the precision of its end-effector \cite{xiao2022concurrently,xiao2020tubular}.
For example, when the tendon-driven robot becomes highly curved to adapt to the environment, the internal friction between the driving tendons and the robot increases significantly \cite{yang2024novel}.
As a result, more complex force-position coupling models are required to achieve accurate control of the end-effector. 
In contrast, fluidically- or magnetically-driven robots offer a simpler approach to control their end effector.
For example, Yuan \textit{et al.} \cite{yuan2024motor} presented a telerobotic endoscope driven by external magnets, enabling independent control of the end effector by an external magnetic field.
Zhang \textit{et al.} \cite{zhang2024pneumaoct} proposed a pneumatic optical coherence tomography (OCT) endoscope.
After reaching deep inside the human body, the distal end of the OCT endoscope could be independently controlled to scan tissue under positive air pressure instead of rotating the whole optic fiber in the proximal end.

Based on the above considerations, a novel pneumatically-driven endoluminal robotic catheter has been proposed in this work.
The robotic catheter has an active bending module with one DoF, which can assist itself in passing through the complex environment.
A novel pneumatic rotatable biopsy mechanism (PRBM) is equipped on the distal end.
In the PRBM, the direction of the biopsy needle can be quickly adjusted under the push of a low-pressure airflow, and the biopsy needle can deployed under the push of a high-pressure airflow, carrying out biopsy, as shown in Fig. \ref{fig:showmodules}.
A comparison between traditional biopsy tools, common continuum endoluminal robots, and the proposed robotic catheter can be seen in the Table. \ref{tab:benchmark} \cite{duan2023novel,wu2022review,lu2023flexible,gao2019continuum}.
The result indicates the superior performance of the proposed robotic catheter in terms of invasiveness, flexibility, and workspace in confined environments.

The contributions of this work are listed as follows: 
\begin{itemize}
    \item A new type of pneumatically-driven endoluminal robotic catheter is proposed along with a novel PRBM with variable biopsy direction omnidirectionally. 
    \item The force analysis and aerodynamic simulation of PRBM have been conducted. 
Analysis of the kinematics model and the biopsy workspace of the proposed robotic catheter have been conducted.
    \item A prototype has been produced.
Repeated deployment assessments of needle deployment have been conducted.
The puncture force of the biopsy needle is measured.
The robotic catheter's ability to perform biopsies after entering tortuous cavities has been demonstrated through phantom tests.
\end{itemize}

%
%
%
%

\section{Design \& Working Principle}

\begin{figure}
    \centering
    \includegraphics[width=0.48\textwidth]{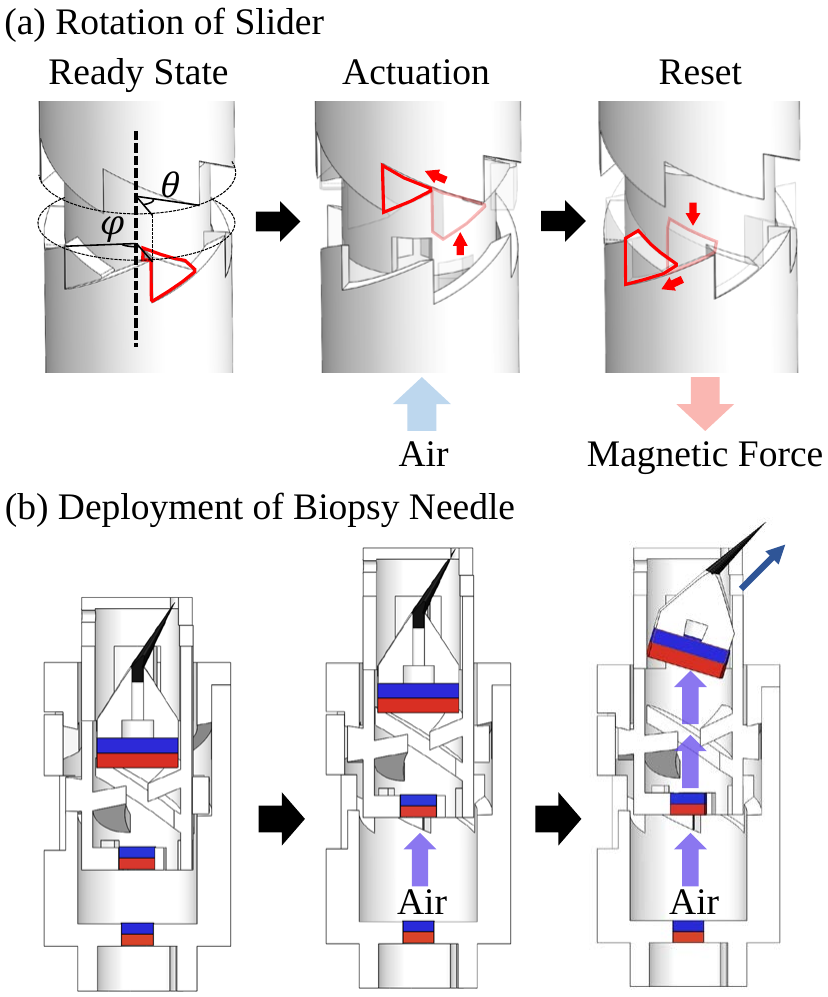}
    \caption{Process of biopsy direction altering and needle deploying.
    (a) Rotation of the slider.
    With a pair of opposite teeth with $\theta$ degree offset.
    The rotational slider has a rotation  $\theta$ after being pushed to match with the upper teeth.
    When the air supply stops, the rotational slider falls to the lower teeth, dragging by magnetic force; then, the slider obtains a rotation of $\varphi$ degree compared to the ready state.
    (b) Deployment of the biopsy needle.
    After increasing the air pressure, the airflow overcomes the magnetic force threshold and finally pushes the slider upward to deploy the needle.}
    \label{fig:rotationandpuncture}
\end{figure}

\subsection{Design of the robotic catheter}
The structure of the proposed robotic catheter is shown in Fig. \ref{fig:design}. 
The catheter consists of three parts: the air-supplying system, the bending module, and the PRBM.
The PRBM comprises a rotation module and a biopsy module. 
The rotation module includes a rotatable slider, a shell cover, and an outer shell.
Both the shell and the shell cover are equipped with a circular array of teeth, with the arrays oriented in opposite directions.
Corresponding triangular protrusions are evenly distributed on the outer wall of the rotatable slider.
Inside the hollow tube of the slider, three fixed fan structures are integrated, generating both propulsive force and rotational torque as the airflow passes through.
The biopsy module includes an upper cover and a slider.
A biopsy needle is mounted tilted on the internal slider, and a hole on the upper cover restricts the direction of needle exertion.
The bending module is fabricated using a combination of two materials with notably different tensile moduli, i.e., FLXA-CT-S50 and FLXA-CT-S85.
Our previous work \cite{zhang2024pneumaoct} has demonstrated the bending module's asymmetric deformation performance under positive pressure. 

In addition to utilizing one air tube to supply positive pressure to the bending module, another air tube is used to regulate the selection of sampling direction and the deployment of the biopsy needle.
Magnets are installed at the bottom of the shell, the rotatable slider, and the slider.
Among them, the magnetic force between the magnets in the shell and the rotatable slider is relatively weak, while the magnetic force between the magnets in the rotatable slider and the slider is comparatively strong.

\subsection{Working Principle of PRBM}
The process of adjusting the biopsy direction can be seen in Fig. \ref{fig:rotationandpuncture}.
During actuation, when the airflow pressure is low, the force exerted by the airflow on the fan structure within the rotatable slider exceeds the magnetic coupling threshold between the rotatable slider and shell, causing the rotatable slider to disengage from the lower teeth on the outer shell. 
Subsequently, the rotatable slider continues to rotate and move upward under the airflow actuation, finally stopping when the triangular bulges engage with the teeth on the upper cover. 
Then, the rotatable slider is rotated by $\theta$ degree in the radial direction relative to the ready state. 
When the airflow input is stopped, the rotatable slider is attracted by the magnetic force and moves downward to engage with the lower teeth.
At this time, the rotatable slider has rotated by $\varphi$ degree in the radial direction relative to the ready state.

When the input airflow pressure is sufficiently high, the airflow not only exceeds the magnetic coupling threshold between the shell and the rotatable slider and pushes the module upward, but also exceeds the magnetic threshold between the rotatable slider and the internal slider, causing the biopsy needle to deploy. 

The PRBM realizes a selective actuation of two functions, the adjustment of the sampling direction and the deployment of the biopsy needle, by using a single air source.
This method not only streamlines the air-driving system but also enhances the structural compactness of the design.

\begin{figure}
    \centering
    \includegraphics[width=0.48\textwidth]{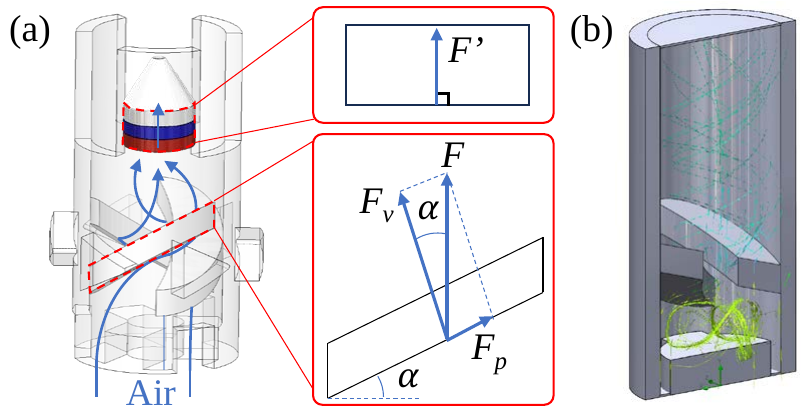}
    \caption{Force analysis and the flow simulation of the rotatable slider.
    (a) When air flows through the array of fixed fans, forces are generated to exert on the fans, which can be decomposed into components perpendicular and parallel to the fans.
    (b) A simulation of the air flows passing through the fixed fans.
    The result also indicates that the airflow would generate force on the fixed fans. 
    }
    \label{fig:airrotate}
\end{figure}
%
%
%
%
\section{Modeling \& Simulation}

\begin{figure}
    \centering
    \includegraphics[width=0.48\textwidth]{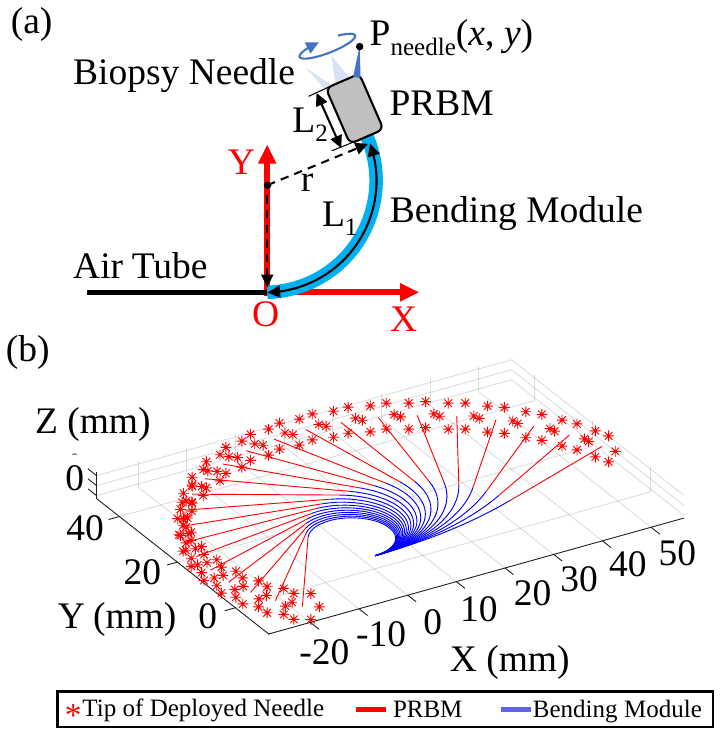}
    \caption{Geometric parameters of the proposed robotic catheter and the simulation of biopsy workspace.
    (a) The shape of the bending module can be estimated with the constant curvature model.
    The position of the distal of the biopsy needle can be obtained.
    (b) The simulation of the biopsy workspace with the module bending, which is obviously larger than the workspace of common biopsy devices.}
    \label{fig:kinematicsmodel}
\end{figure}

\begin{figure}[t]
    \centering
    \includegraphics[width=0.48\textwidth]{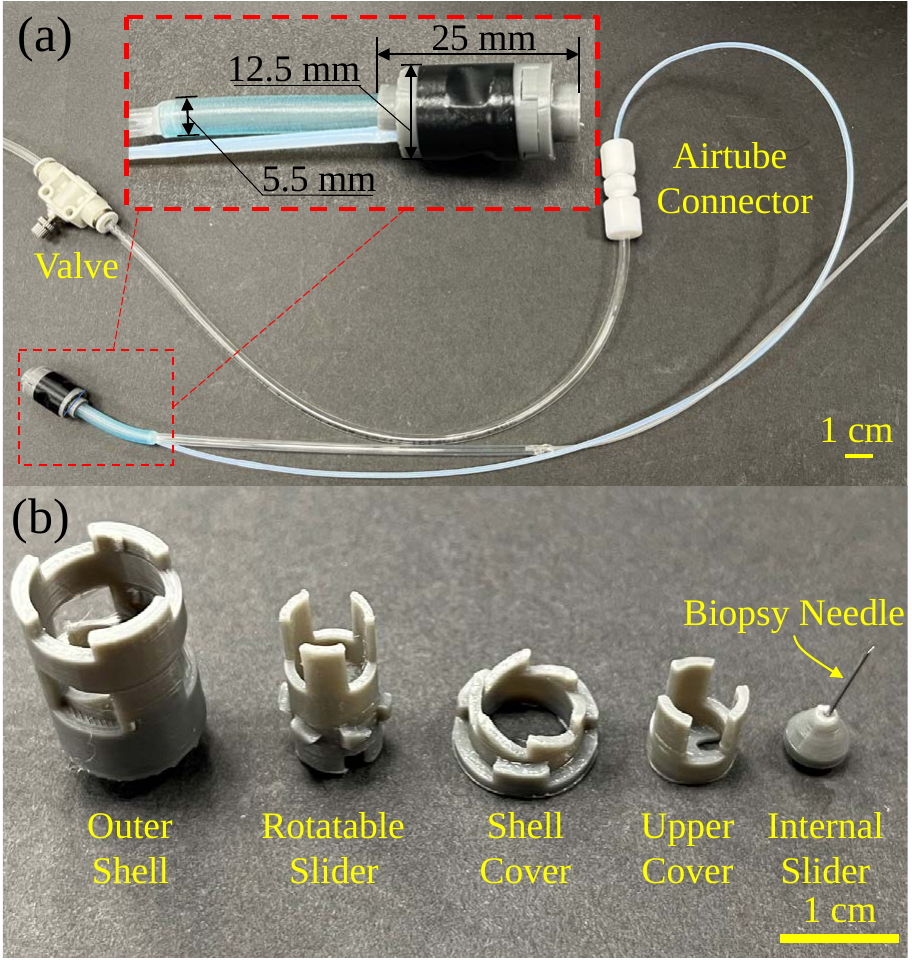}
    \caption{Overview of the robotic catheter prototype and its components.
    (a) The outer diameter and length of the PRBM prototype are 12.5 $mm$ and 25 $mm$.
    The outer diameter of the used bending module is 5.5 $mm$.
    With a pump supplying air, the robotic catheter prototype can be simply controlled by a valve.
    (b) The components of the PRBM are made by 3D printing with PLA material.}
    \label{fig:prototype}
\end{figure}
\subsection{Kinematics Model}
In previous work \cite{zhang2024pneumaoct}, an analysis of the characteristics of the bending module was conducted, revealing the relationship between the magnitude of positive pressure and the radius of the bending module. 
When the positive air pressure is applied, the module will bend with an approximately constant curvature. 
Since the radial and axial expansions are small relative to the entire length of the module, they are ignored in this kinematic modeling.
The geometric relationship of the proposed robotic catheter is shown in Fig. \ref{fig:kinematicsmodel}(a).
According to the constant curvature assumption, the posture of the catheter can be derived as follows.
\begin{align}
    \theta r &= L_1\,,\\
    P_{needle} &= \textbf{Rot}(\theta)\textbf{T}(L_2)P_{origin}\,,
\end{align}
where the $r$, $\theta$, and $L_1$ denote the radius, bending angle, and length of the bending module, respectively, and the $\textbf{Rot}(\theta)$ denotes the rotation matrix about distal of the bending module respect to the center about $\theta$, the $\textbf{T}(L_2)$ denotes the translation matrix along the direction of distal of bending module with distance of $L_2$.

According to the forward kinematic model, we can simulate the biopsy workspace of the proposed robotic catheter, and the simulation results are shown in Fig. \ref{fig:kinematicsmodel}(b).
With six variable biopsy directions at each distal position, the proposed robotic catheter significantly expands the biopsy workspace compared to traditional biopsy tools and common continuum endoluminal robots.

\subsection{Internal Force Model}
During the working process, the airflow passing through the PRBM can generate different motions of the components.
Alternating the air pressure enables a switch between the actions of alternating biopsy direction and deploying the biopsy needle.
The force generated on one fan structure can be analyzed, as shown in Fig. \ref{fig:kinematicsmodel}(a), and the equation can be derived as follows.
\begin{align}
    \textbf{F} = Pa = \textbf{F}_p+\textbf{F}_v\,
    \label{eq:F=pa}
\end{align}
where $P$ denotes the air pressure, the $a$ denotes the area of a single fan structure, $\textbf{F}$ denotes the force generated on the fan structure by the airflow, and $\textbf{F}_p$ and $\textbf{F}_v$ denote the division of $\textbf{F}$ which are parallel and vertical, respectively. 
The force pushing the internal slider to move upward can also use equation (\ref{eq:F=pa}) to derive, while the $a$ becomes the area of the internal slider's bottom surface.
Additionally, a flow simulation of the air passing through the rotatable slider has been carried out, as shown in Fig. \ref{fig:kinematicsmodel}(b).
The particle flow trajectory indicates that the airflow passing through the array of fan structures generates forces to push and rotate the rotatable slider, consistent with the analysis.

The forces between the magnets inside the PRBM can be calculated by using the integration method.
\begin{align}
    \textbf{F}=\frac{\mu_0}{4\pi}\int_{V_1}
    \int_{V_2}\frac{3(\textbf{M}_1\cdot\hat{\textbf{r}})(\textbf{M}_2\cdot\hat{\textbf{r}})-\textbf{M}_1\cdot\textbf{M}_2}{r^5}\textbf{r}dV_1dV_2\,,
\end{align}
where $\textbf{M}$ denotes the magnetization of the magnet, $\textbf{r}$ denotes the relative position vector between two magnets, $\hat{(\cdot)}$ denotes the unit vector, $r$ denotes the modulus of $\textbf{r}$, $V$ denotes the volume of the magnet, and $\mu_0$ denotes the magnetic permeability of vacuum.

By quantifying the air-pushing forces within the mechanism and the magnetic attraction force, the performance of the PRBM can be improved by optimizing the shape of the fan structure and the type of magnet used.

%
%
%
%
\section{Prototype \& Experiment}

\begin{figure}[t]
    \centering
    \includegraphics[width=0.48\textwidth]{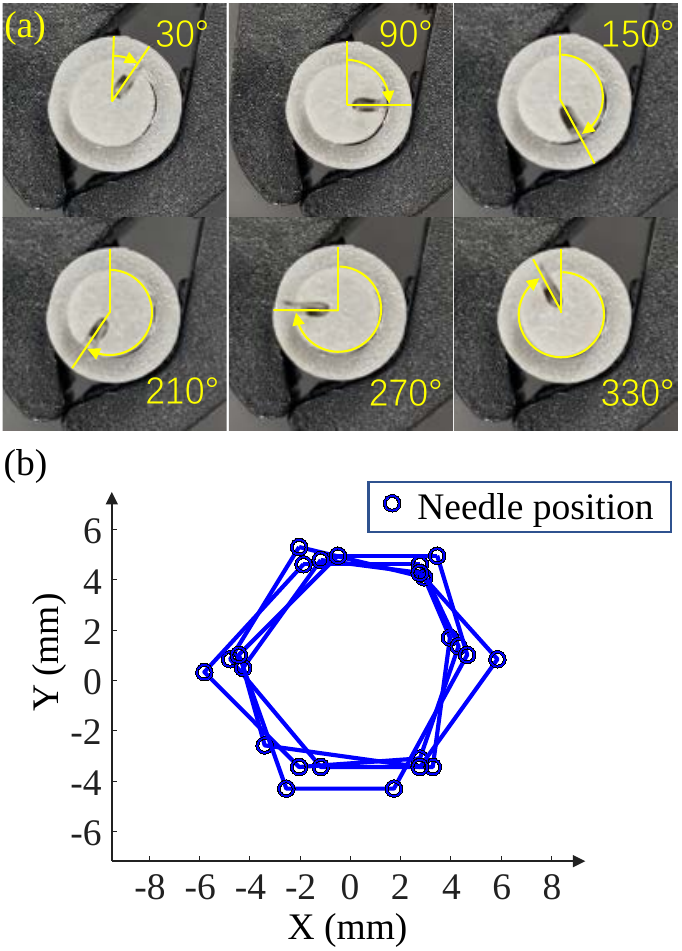}
    \caption{Repeated deployment assessment of the biopsy needle.
    (a) In the current design, the biopsy needle can be deployed in six different directions, with a 60-degree interval between adjacent directions.
    (b) A record of four consecutive rounds of needle redeployment.}
    \label{fig:needleposition}
\end{figure}

\begin{figure}
    \centering
    \includegraphics[width=0.48\textwidth]{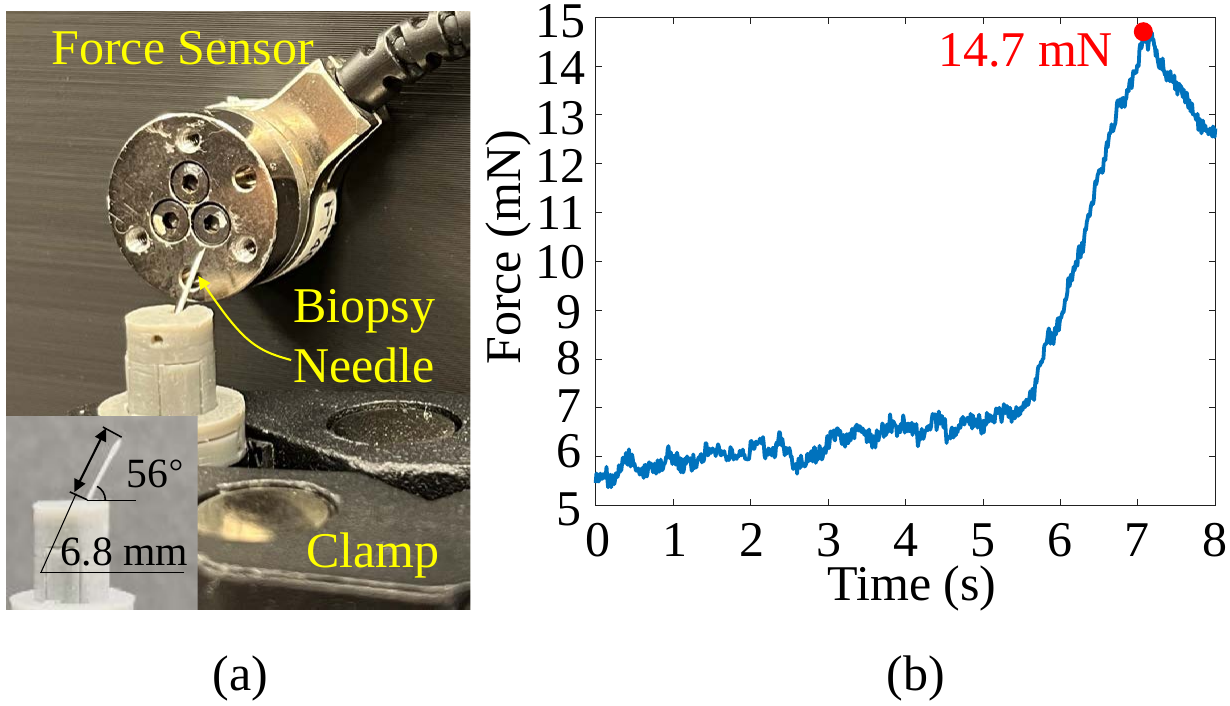}
    \caption{Collecting the puncture force of the biopsy needle.
    (a) Setup of the force measuring.
    The PRBM is held by a clamp, and a force sensor is used to collect the puncture force of the needle.
    The tilted angle of the extended biopsy needle is $56$ degrees, and the length of the extended part is 6.8 $mm$.
    (b) The record data of the force sensor. 
    As the result shows, the puncture force is about 10 $mN$ when the supplied air pressure is about 0.3 $MPa$.}
    \label{fig:forcesensing}
\end{figure}


\begin{figure}
    \centering
    \includegraphics[width=0.45\textwidth]{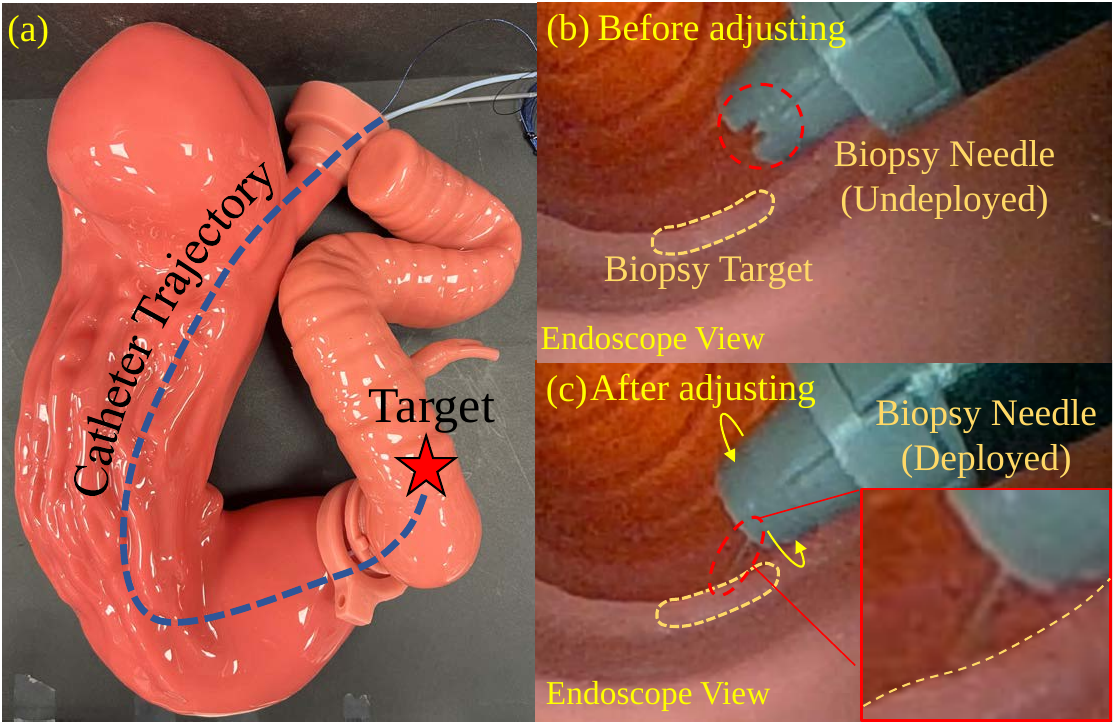}
    \caption{Phantom test of the robotic catheter prototype.
    (a) A stomach-duodenum phantom was used.
    The red star denotes the biopsy target on the duodenum wall.
    The blue dashed line denotes the occluded catheter trajectory.
    (b) Upon initial insertion, the working direction of the biopsy needle was not aligned with the target.
    (c) After actuating the slider to rotate, the biopsy needle was finally aligned with the target, and then the needle was deployed.}
    \label{fig:phantomtest}
\end{figure}
\subsection{Prototype Design}
A prototype of the proposed pneumatically-driven robotic catheter has been manufactured, as shown in Fig. \ref{fig:prototype}.
The maximum outer diameter and length of the PRBM are 12.5 $mm$ and 25 $mm$, respectively.
The outer diameter of the used bending module is 5.5 $mm$.
The sizes of magnets used in the outer shell, rotatable slider, and internal slider are $1mm \times 2.5mm$ (length$\times$diameter), $2mm \times 3mm$, and $1mm \times 3mm$, respectively. 
The material of the used magnets is N52.
After testing and optimizing, the corresponding air pressures used to adjust the biopsy direction and deploy the biopsy needle were chosen to be 0.2 $MPa$ and 0.3 $MPa$.

In the prototype, the PRBM has six different puncture directions, as shown in Fig. \ref{fig:needleposition}(a), and there is a 60-degree interval between the adjacent directions.
Using pneumatic actuation, the adjustment of direction and deployment of the needle can ideally be completed in approximately 1 second.

\subsection{Characterization and Demonstration}
First, repeated deployment assessments of the needle were conducted.
Four rounds of repeated needle deploying were carried out, and the positions of the needle tip were recorded by a camera.
The result can be seen in Fig. \ref{fig:needleposition}(b).
The average position deviation of the deployed needle tip in the plane is 2.63 $mm$.

The puncture force exerted by the biopsy needle under a 0.3 $MPa$ actuation air pressure was measured.
The force measurement setup is shown in Fig. \ref{fig:forcesensing}, where a six-axis force sensor (Nano17-E Transducer, ATI Industrial Automation) is used to collect the force generated by the needle.
After applying a moving average filter with a window of size 19, most noise of the collecting data is filtered out.
The result can be seen in Fig. \ref{fig:forcesensing}, and the puncture force is measured at approximately 10 $mN$, which is sufficient for the biopsy needle to sample the soft human tissue.

Phantom tests have been conducted to demonstrate the feasibility of the proposed robotic catheter to adjust puncture omnidirectionally \textit{in situ}, as shown in Fig. \ref{fig:phantomtest}.
A stomach-duodenum phantom was utilized, with the biopsy target estimated to be located on the duodenal wall.
An endoscope was inserted alongside the robotic catheter prototype to locate the target.
Upon initial insertion, although the robotic catheter successfully reached the biopsy site, its biopsy needle was not aligned with the target, as shown in Fig. \ref{fig:phantomtest}(b).
After actuating the mechanism to rotate, the puncture direction was adjusted toward the target.
The biopsy needle was then deployed by increasing the input air pressure and successfully puncturing the target.
The entire adjustment process was completed within 10 seconds, minimizing relative movement between the catheter's body and its surrounding environment while swiftly and accurately adjusting the biopsy needle's puncturing direction.
The test demonstrated the feasibility of the proposed design for performing omnidirectional biopsies deep inside the human body.

%
%
%
%
\section{Conclusion}
 \begin{figure}[t]
     \centering\includegraphics[width=0.35\textwidth]{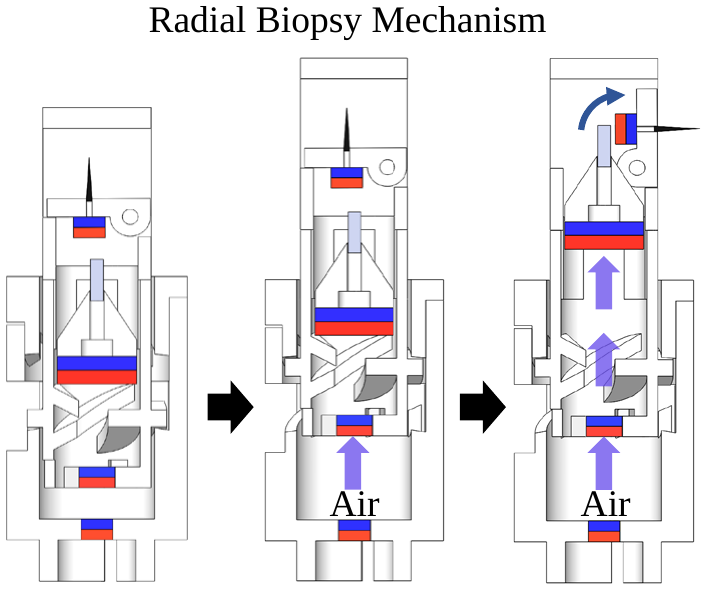}
     \caption{Design of a possible modular PRBM that can carry out a radial biopsy.}
     \label{fig:radialbiopsymechanism}
 \end{figure}

This study introduces a new pneumatically operated endoluminal robotic catheter with PRBM. 
The PRBM enables adjusting the biopsy direction and deploying the needle omnidirectionally by using a single air source.
With the design, the user can adjust the working direction of the PRBM quickly and precisely after the robotic catheter advances deep inside the human body, and the relative motion between the catheter's body and the surrounding environment is minimized.
Compared with other biopsy devices, the proposed robotic catheter demonstrates its potential to perform biopsies quickly, safely, and accurately deep inside the human body, providing new ideas for the design of biopsy devices.

Nonetheless, there are numerous elements of the robotic catheter that could be enhanced. The configuration of the PRBM can be refined to achieve improved aerodynamic efficiency, resulting in a stronger, more stable and precise biopsy. Additionally, the dimensions of the PRBM could be minimized to suit a wider range of application scenarios. Furthermore, the possibilities of the modular design of the PRBM can be investigated.
For example, one possible design can be seen in Fig. \ref{fig:radialbiopsymechanism}, which can perform a radial biopsy by replacing two components.
In the future, by replacing modular components, the PRBM can quickly switch between different biopsy modes, realize customized biopsy for patients, and make the biopsy process easier and safer.

\addtolength{\textheight}{-12cm}   







%
%
%
%
\bibliographystyle{IEEEtran}

\bibliography{ref}

\begin{thebibliography}{10}
\providecommand{\url}[1]{#1}
\csname url@samestyle\endcsname
\providecommand{\newblock}{\relax}
\providecommand{\bibinfo}[2]{#2}
\providecommand{\BIBentrySTDinterwordspacing}{\spaceskip=0pt\relax}
\providecommand{\BIBentryALTinterwordstretchfactor}{4}
\providecommand{\BIBentryALTinterwordspacing}{\spaceskip=\fontdimen2\font plus
\BIBentryALTinterwordstretchfactor\fontdimen3\font minus \fontdimen4\font\relax}
\providecommand{\BIBforeignlanguage}[2]{{%
\expandafter\ifx\csname l@#1\endcsname\relax
\typeout{** WARNING: IEEEtran.bst: No hyphenation pattern has been}%
\typeout{** loaded for the language `#1'. Using the pattern for}%
\typeout{** the default language instead.}%
\else
\language=\csname l@#1\endcsname
\fi
#2}}
\providecommand{\BIBdecl}{\relax}
\BIBdecl

\bibitem{strnad2024percutaneous}
B.~S. Strnad, M.~Kristeva, M.~Itani, D.~T. Fetzer, S.~D. O'Connor, M.~D. Patel, and W.~D. Middleton, ``Percutaneous core biopsy devices: A detailed review and comparison of different needle designs,'' \emph{Ultrasound Quarterly}, vol.~40, no.~1, pp. 1--19, 2024.

\bibitem{lin2022modular}
B.~Lin, J.~Wang, S.~Song, B.~Li, and M.~Q.-H. Meng, ``A modular lockable mechanism for tendon-driven robots: Design, modeling and characterization,'' \emph{IEEE Robotics and Automation Letters}, vol.~7, no.~2, pp. 2023--2030, 2022.

\bibitem{duan2024survey}
Y.~Duan, J.~Ling, Z.~Feng, T.~Ye, T.~Sun, and Y.~Zhu, ``A survey of needle steering approaches in minimally invasive surgery,'' \emph{Annals of Biomedical Engineering}, vol.~52, no.~6, pp. 1492--1517, 2024.

\bibitem{dupourque2019transbronchial}
L.~Dupourqu{\'e}, F.~Masaki, Y.~L. Colson, T.~Kato, and N.~Hata, ``Transbronchial biopsy catheter enhanced by a multisection continuum robot with follow-the-leader motion,'' \emph{International journal of computer assisted radiology and surgery}, vol.~14, no.~11, pp. 2021--2029, 2019.

\bibitem{wu2017development}
L.~Wu, S.~Song, K.~Wu, C.~M. Lim, and H.~Ren, ``Development of a compact continuum tubular robotic system for nasopharyngeal biopsy,'' \emph{Medical \& biological engineering \& computing}, vol.~55, pp. 403--417, 2017.

\bibitem{pittiglio2022patient}
G.~Pittiglio, P.~Lloyd, T.~da~Veiga, O.~Onaizah, C.~Pompili, J.~H. Chandler, and P.~Valdastri, ``Patient-specific magnetic catheters for atraumatic autonomous endoscopy,'' \emph{Soft robotics}, vol.~9, no.~6, pp. 1120--1133, 2022.

\bibitem{russo2023continuum}
M.~Russo, S.~M.~H. Sadati, X.~Dong, A.~Mohammad, I.~D. Walker, C.~Bergeles, K.~Xu, and D.~A. Axinte, ``Continuum robots: An overview,'' \emph{Advanced Intelligent Systems}, vol.~5, no.~5, p. 2200367, 2023.

\bibitem{li2023three}
L.~Li, X.~Li, B.~Ouyang, H.~Mo, H.~Ren, and S.~Yang, ``Three-dimensional collision avoidance method for robot-assisted minimally invasive surgery,'' \emph{Cyborg and Bionic Systems}, vol.~4, p. 0042, 2023.

\bibitem{xiao2022concurrently}
X.~Xiao, Y.~Wu, Q.~Wu, and H.~Ren, ``Concurrently bendable and rotatable continuum tubular robot for omnidirectional multi-core transurethral prostate biopsy,'' \emph{Medical \& Biological Engineering \& Computing}, vol.~60, pp. 229--238, 2022.

\bibitem{xiao2020tubular}
X.~Xiao, C.~Li, X.~Gu, Y.~Yan, Y.~Wu, Q.~Wu, E.~Chiong, and H.~Ren, ``A tubular dual-roller bending mechanism toward robotic transurethral prostate biopsy,'' \emph{IEEE/ASME Transactions on Mechatronics}, vol.~26, no.~5, pp. 2483--2494, 2020.

\bibitem{yang2024novel}
Z.~Yang, L.~Yang, Y.~Sun, and X.~Chen, ``A novel contact-aided continuum robotic system: design, modeling, and validation,'' \emph{IEEE Transactions on Robotics}, 2024.

\bibitem{yuan2024motor}
S.~Yuan, C.~Xu, B.~Cui, T.~Zhang, B.~Liang, W.~Yuan, and H.~Ren, ``Motor-free telerobotic endomicroscopy for steerable and programmable imaging in complex curved and localized areas,'' \emph{Nature Communications}, vol.~15, no.~1, p. 7680, 2024.

\bibitem{zhang2024pneumaoct}
T.~Zhang, S.~Yuan, C.~Xu, P.~Liu, H.-C. Chang, S.~H.~C. Ng, H.~Ren, and W.~Yuan, ``Pneumaoct: Pneumatic optical coherence tomography endoscopy for targeted distortion-free imaging in tortuous and narrow internal lumens,'' \emph{Science Advances}, vol.~10, no.~35, p. eadp3145, 2024.

\bibitem{duan2023novel}
X.~Duan, D.~Xie, R.~Zhang, X.~Li, J.~Sun, C.~Qian, X.~Song, and C.~Li, ``A novel robotic bronchoscope system for navigation and biopsy of pulmonary lesions,'' \emph{Cyborg and Bionic Systems}, vol.~4, p. 0013, 2023.

\bibitem{wu2022review}
K.~Wu, B.~Li, Y.~Zhang, and X.~Dai, ``Review of research on path planning and control methods of flexible steerable needle puncture robot,'' \emph{Computer Assisted Surgery}, vol.~27, no.~1, pp. 91--112, 2022.

\bibitem{lu2023flexible}
M.~Lu, Y.~Zhang, C.~M. Lim, and H.~Ren, ``Flexible needle steering with tethered and untethered actuation: Current states, targeting errors, challenges and opportunities,'' \emph{Annals of Biomedical Engineering}, vol.~51, no.~5, pp. 905--924, 2023.

\bibitem{gao2019continuum}
Y.~Gao, K.~Takagi, T.~Kato, N.~Shono, and N.~Hata, ``Continuum robot with follow-the-leader motion for endoscopic third ventriculostomy and tumor biopsy,'' \emph{IEEE Transactions on Biomedical Engineering}, vol.~67, no.~2, pp. 379--390, 2019.

\end{thebibliography}

\end{document}